\documentclass{article}
\usepackage{spconf,amsmath,graphicx}
\usepackage{preamble}

\usepackage[belowskip=-10pt,aboveskip=5pt]{caption}
\newcommand\blfootnote[1]{%
  \begingroup
  \renewcommand\thefootnote{}\footnote{#1}%
  \addtocounter{footnote}{-1}%
  \endgroup
}

\title{Multi-Head ReLU Implicit Neural Representation Networks}
\name{Arya~Aftab$^{1,2}$, Alireza~Morsali$^{3}$, Shahrokh Ghaemmaghami$^{1,2}$
}
\address{$^{1}$ 
Department of Electrical Engineering, Sharif University of Technology, Tehran, Iran
\\$^{2}$
Electronics Research Institute, Sharif University of Technology, Tehran, Iran
\\$^{3}$ 
Department of Electrical and Computer Engineering, McGill University, Montreal, Canada\\
\normalsize \texttt{aftab.arya@ee.sharif.edu,  alireza.morsali@mail.mcgill.ca, ghaemmag@sharif.edu}}
\begin{document}
%\ninept
%
\maketitle
\begin{abstract}
In this paper, a novel multi-head multi-layer perceptron (MLP) structure is presented for implicit neural representation (INR). Since conventional rectified linear unit (ReLU) networks are shown to exhibit spectral bias towards learning low-frequency features of the signal, we aim at mitigating this defect by taking advantage of local structure of the signals. To be more specific, an MLP is used to capture the global features of the underlying generator function of the desired signal. Then, several heads are utilized to reconstruct disjoint local features of the signal, and to reduce the computational complexity, sparse layers are deployed for attaching heads to the body. Through various experiments, we show that the proposed model does not suffer from the special bias of conventional ReLU networks and has superior generalization capabilities. Finally, simulation results confirm that the proposed multi-head structure outperforms existing INR methods with considerably less computational cost. The source code is available at {\color{purple}\url{https://github.com/AlirezaMorsali/MH-RELU-INR}}
\end{abstract}
\begin{keywords}
Implicit neural representation, multi-head MLP, ReLU network, spectral bias, multi-layer perceptron
\end{keywords}
\section{Introduction}
\label{sec:intro}
Recently, there has been a considerable interest in implicit neural representation (INR) for parameterizing various kinds of signals \cite{park_2019,Fourier_features, sitzmann2020implicit, mildenhall2020nerf}. For example, images and audio signals are conventionally stored as discrete grids of pixels and discrete samples of amplitudes, respectively \cite{yan2016perspective, lombardi2019neural}. On the other hand, with INR, the goal is to find a continuous generator function for the target signal\cite{meshry2019neural, aliev2020neural, thies2019deferred, liu2019neural,Sal1,Modulated,DeepVoxels,peng2020convolutional}. In particular, for images as the target signal of INR, the pixel coordinates are mapped to RGB color values.\blfootnote{This work was supported by DGSculptor (\url{www.dgsculptor.com}).}

%A 3D shape can also be parameterized with voxel grids \cite{yan2016perspective, lombardi2019neural}, point clouds \cite{meshry2019neural, aliev2020neural}, or meshes \cite{thies2019deferred, liu2019neural}.

Multi-layer perceptron (MLP) networks and, in particular, deep neural networks (DNN) are shown to have an unprecedented capability in learning various input-output transformations (even random input-output mappings) with high accuracy \cite{arpit2017closer, zhang2021understanding}. One of the widely used activation functions in DNNs is rectified linear unit (ReLU), which brings considerable advantages, such as overcoming gradient vanishing and facilitating the learning process. Furthermore, it is proven that ReLU neural networks can express any continuous piece-wise linear function, which forms bounded convex polytopes. Moreover, every continuous piece-wise linear function can be defined by a specific ReLU neural network \cite{arora2018optimization, arora2016understanding, hanin2019universal}. As a result, ReLU networks are known as universal approximators.
Consequently, ReLU networks could be the prime candidate for INR.
However, in \cite{rahaman2019spectral}, the Fourier transform of ReLU neural networks reveals a spectral bias in such networks, i.e., ReLU networks tend to learn lower frequencies faster. 

It is shown that an increase in the complexity of the data manifold shapes and low data dimension eases the learning of the high frequencies \cite{Fourier_features}.
Therefore, authors in \cite{Fourier_features} proposed manipulating the input data by a sine kernel method to facilitate learning high-frequency signals. Positional encoding relieves the spectral bias by kernel regression module, which maps a low-dimension input to a high-dimension one through a set of sine and cosine functions with different frequencies.
Furthermore, the use of periodic activation functions for INR is presented in \cite{sitzmann2020implicit}, which allows the model (namely, SIREN) to learn high-frequency data more effectively.
Nevertheless, a major drawback of the aforementioned methods is the computational cost of such networks during both training and inference. Moreover, in case of SIREN, due to the use of periodic activation function, the network is highly sensitive to initialization and can exhibit unstable behaviours for different types of signals.

% \subsection{Capacitating DNNs to learn higher frequency data} MLP networks have been exploited in
%

% numerous computer vision and graphics tasks. Representing various signals including images and 3D scenes in discrete domains has several limitations in sampling and storing data points. Thus, the attentions were attracted toward utilizing MLPs in representation of these signals in continuous domain. Nevertheless, ReLU MLPs defectively learn high-frequency signals like images. Previous works have mitigated this issue, i.e., spectral bias, using different approaches. In \cite{tancik2020fourier, mildenhall2020nerf}, positional encoding method has been exploited to generate high frequency outputs.   \fxnote{add 7-8 references}

% Conventional multi-layer perceptron (MLP) networks or deep neural networks (DNNs) with rectified
% linear unit (ReLU) activation functions are partial to learn low-frequency data faster (also known
% as spectral bias \cite{rahaman2019spectral}) which is not ideal for INR tasks.

In this paper, we present an efficient multi-head INR network structure which is capable of learning high-frequency signals accurately with considerably lower computational cost compared to existing methods. The target signals of INR, for instance image and audio, usually exhibit local structure and neighbourhood dependencies. By exploiting this property of the target signals, we show that ReLU networks can be boosted to be less partial to low-frequency components during the training phase. We introduce a multi-head network architecture in which the body learns the global features of the signal and the output layers, consists of several heads, reconstruct separate parts of the signal, and learn local features of the signal. This approach has several promising advantages: I. The network architecture can be tailored to learn high-frequency components without incorporating periodic functions. II. The number of floating point operations (FLOPs) required to generate a signal decreases significantly as the number of heads of the model increases. III. The generalization ability of the model intensifies in the multi-head models compared to the base model.

% One of the remarkable advantage of regression models is generalization of the input space.
% Therefore, We investigate
% Considering that these networks are continuous functions for generating the trained signal, their generalization power to learn the 

% In this work, we propose a Multi-head SIREN method, which generates more than 

% \section{Related Works}
% \label{sec:related}
% \subsection{Spectral Bias of Neural Networks}
\begin{figure*}[t]
\vspace{-3em}
			\centering
			\includegraphics[width=165mm,scale=1.0]{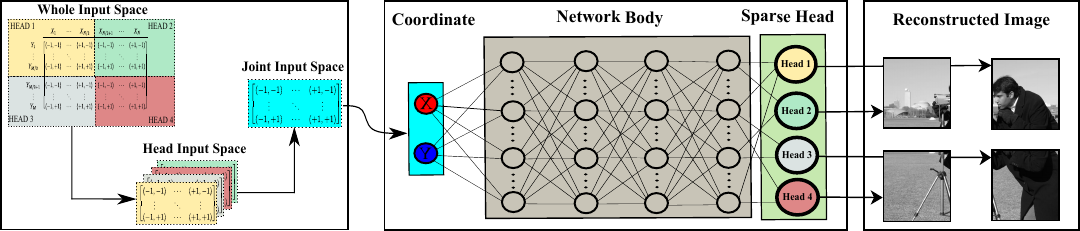}
				\caption{\small Network Architecture for 4 Heads}
				\label{fig:architecture}
				
			\vspace{-1mm}
			\end{figure*}

\section{Problem Formulation}
In this work, we focus on grayscale images for INR. However, the proposed method can be generalized to other multi-dimensional signals.
For a given $N_x\times N_y$ grayscale image, i.e., $\I\in\mathbb{R}^{N_x\times N_y}$, the goal of INR is to find an underlying function $\Phi: \mathbb{R}^2\to\mathbb{R}$ that maps pixel coordinates $r$ and $c$ into the pixel value $\I[r, c]$ for $r=1,2,\dots,N_x$ and $c=1,2,\dots,N_y$ where $\I[r, c]$ denote the grayscale value of the $\I$ on the $r$th row and the $c$th column.
Since the function $\Phi$ is continuous, we only have its values at discrete points, i.e.,
\begin{equation}
  \label{eq:1}
     \Phi\big{(}x(r), y(c)\big{)} = \I[r, c]
\end{equation}
where $x(r)=2\frac{r-1}{N_x-1}-1$, $y(c)=2\frac{c-1}{N_y-1}-1$. Note that any arbitrary interval can be selected as the domain of the function; however, for simplicity $x,y\in[-1,1]$ is usually chosen.

Existing works in the literature \cite{sitzmann2020implicit,Fourier_features} use an MLP to approximate $\Phi(x, y)$ and train the MLP using the pixel coordinate and the corresponding pixel values as the dataset:
\[
  \pazocal{D}=\bigg\{\Big(\big(x(r), y(c)\big), \I[r, c]\Big)\bigg\}_{r,c=1}^{N_x, N_y}.
\]
% However, in Equation \eqref{eq:1}, the $\Phi$ function is a generic mapping that we can be further expressed as the sum of $M$ sub-functions to reduce computational complexity.
% In particular we can write
% \begin{equation}
%   \label{eq:2}
%   \Phi(x_1, x_2)=\sum_{m=1}^M\phi_m(x_1, x_2)I_m(x_1, x_2)
% \end{equation}
% Where $\phi_n(x_1, x_2)$'s are continuous functions from $\mathbb{R}^2$ to $\mathbb{R}$ and $I_n(x_1, x_2)$'s are indicator functions.

\section{Proposed Method}
In order to alleviate the spectral bias of the ReLU networks, we take advantage of the local structure of the target signal by dividing the input domain $x$ and $y$ into ${H_x}$ and  ${H_y}$ equal intervals, respectively, where $H_x$ and $H_y$ are non-negative integers. Let us further assume $N_x$ and $N_y$ are divisible by $H_x$ and $H_y$, respectively.
% To be more specific, by defining
% \begin{equation}
%   \label{eq:3}
%   I_m=[]
% \end{equation}
Consequently, the image $\I$ can be divided into $M = {H_x}{H_y}$ equal grid cells, each of which is of size $\hat{N}_x \times \hat{N}_y$ where $\hat{N}_x=\frac{N_x}{{H_x}}$ and $\hat{N}_y=\frac{N_y}{{H_y}}$. Therefore, for $\hat{r}=1,2,\dots,\hat{N}_x$ and $\hat{c}=1,2,\dots,\hat{N}_y$, we can further explicitly write the image cells as
\begin{equation}
  \label{eq:4}
  \I_{l,k}[\hat{r},\hat{c}]= \I[\hat{N}_h(l-1)+\hat{r}, \hat{N}_w(k-1)+\hat{c}]
\end{equation}
where $l=1,2,\dots,{H_x}$ and $k=1,2,\dots,{H_y}$.

Consequently, the INR of a given image $\I$ is broken down into finding $M$ functions. Thus, instead of having a generator function $\Phi(x,y)$ for the whole image $\I$, we use $M$ functions $\phi_{l,k}(\hat{x},\hat{y})$ for the corresponding cell of the image $\I_{l,k}$, i.e.,
\begin{equation}
  \label{eq:11}
     \phi_{l,k}\big{(}\hat{x}(\hat{r}), \hat{y}(\hat{c})\big{)} = \I_{l,k}[\hat{r}, \hat{c}]
\end{equation}
where $\hat{x}(\hat{r})=2((\hat{r}-1)/N_x)-1$, $\hat{y}(\hat{c})=2((\hat{c}-1)/N_y)-1$.
% If we approximate each image slice by a $\phi$ function, we need $M$ functions to approximate the whole image.

However, training $M$ MLPs is not efficient. Moreover, the desired signals (images in this case) usually contain global features which can be shared by all $M$ functions. Motivated by this fact, we use a function (as body) to produce global features denoted by $\psi({\hat{x}, \hat{y}})$. This function, in essence, behaves as an embedding that maps coordinates to a high-dimensional space.
Subsequently, $M$ disjoint rendering functions $\tau_{l,k}(\cdot)$ use these mapped coordinates to reconstruct the details of each cell of the image. We can therefore write:
\begin{equation}
    \label{eq:2}
    \phi_{l,k}(\hat{x},\hat{y}) = \tau_{l,k}(\psi(\hat{x},\hat{y})).
\end{equation}

% Since the size of the image slices is equal, we can consider the input coordinates to be the same. Due to the equality of the input and assumption that the $\psi$ functions is only responsible for embedding, this part of the $\phi$ functions can be considered the same.
% To solve Equation \eqref{eq:1}, we offer it in the form of an optimization problem. According to Equation \eqref{eq:1}, \eqref{eq:2}, and the noted points, the objective function can be presented as Equation \eqref{eq:objective_function}.

To find such functions via training, we can write the following optimization problem:
\begin{equation}
    \label{eq:objective_function}
    \min_{\tau_{l,k}(),\psi()} \sum_{\hat{r}=1}^{\hat{N}_h}\sum_{\hat{c}=1}^{\hat{N}_w}\sum_{l=1}^{H_x}\sum_{k=1}^{H_y}\bigg(\tau_{l,k}\Big(\psi\big(\hat{x}(\hat{r}), \hat{y}(\hat{c})\big)\Big) - \I_{l,k}[\hat{r}, \hat{c}]\bigg)^2.
  \end{equation}

% Where $x(\hat{r})=2(\hat{r}/{\hat{N}_h})-1$, $y(\hat{c})=2(\hat{c}/{\hat{N}_w})-1$, and $s_m[\hat{r}, \hat{c}]$ denote the value of the $m$th slice of the image on the $\hat{r}$th row and the $\hat{c}$th column.
% \begin{equation}
%     s_m[r,c] = S[((m-1)*2^{H_1})+r,((m-1)*2^{H_2})+c]
% \end{equation}
\subsection{Network Architecture}
Here, we take advantage of expressive power of MLPs to approximate the rendering functions $\tau_{l,k}(\cdot)$ as well as the body function $\psi(\hat{x},\hat{y})$. In particular, we propose a multi-head implicit neural representation network consisting of two parts, namely, the body and the rendering heads. The body is a regular MLP with ReLU activation function. While, the rendering networks are a special sparse layer with multiple outputs. Fig. \ref{fig:architecture} illustrates the structure of the proposed network, which is explained below in details.
%A $4$-layer MLP named the body is collectively used for parts $\psi$ of $\phi$ functions.
%One neuron is used for each $\tau$ part to approximate the pixel values of each slice of the image.
%The set of these single neurons can be thought of as an fully connected layer called the rendering multi-head part, which is attached to the model body.
%The main problem of the model's rendering multi-head part can be considered the high number of parameters, which is proportional to the number of heads connected to the model body.
%The significant number of parameters required by the proposed structure occurs when the number of heads converges to the number of pixels in the image.
%For example, if $256^2$ heads are used for a $256 \times 256$ grayscale image and the last layer of the model body has $256$ neurons, then approximately $16.7$ million parameters are required.
%To solve the problem, we proposed a sparse layer instead of the fully connected layer. In a sparse layer, the number of trainable parameters is significantly reduced due to removing some connections with the previous layer.
%Since each neuron reconstructs a specific part of the image, it must connect to the previous layer. We denote the number of connections of each neuron with the previous layer by the parameter $\alpha$.
%We consider the minimum value of parameter $\alpha$ to be $1$, and by increasing this parameter to the number of neurons in the previous layer, the sparse layer becomes a fully connected layer. 

\begin{figure*}[t]
\vspace{-1em}
			\centering
			\includegraphics[width=155mm,scale=1.0]{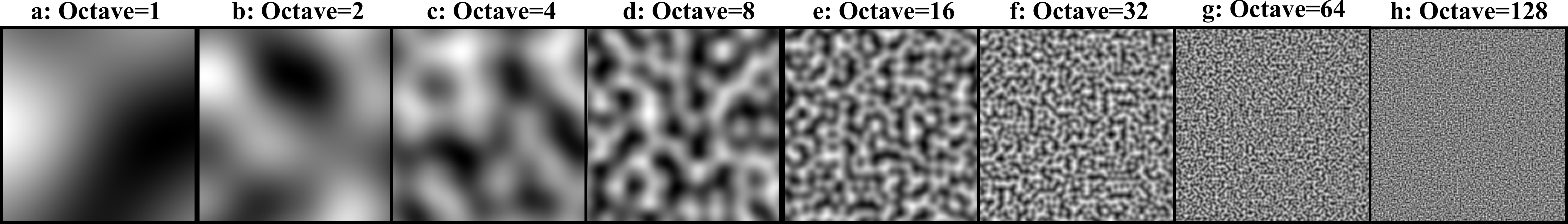}
				\caption{\small Perlin noise}
				\label{fig:perlin_niose}
			\end{figure*}

\subsubsection{Body}
A $4$-layer MLP with ReLU activation function is used as the body to approximate $\psi(\hat{x},\hat{y})$. This MLP takes the normalized coordinates of a pixel and creates an intermediate vector that embeds the coordinate into high-dimensional space that is fed into the rendering networks.
%The body is comprised of a 4-layer MLP with ReLU activation function.  This MLP takes the normalized coordinates of a pixel and creates an intermediate vector that encapsulates the information that is fed into the next stage. Further, the vector is connected to the rendering generating pixel values.
\begin{figure}[t]
% \vspace{-1em}
			\centering
			\includegraphics[scale = 0.57]{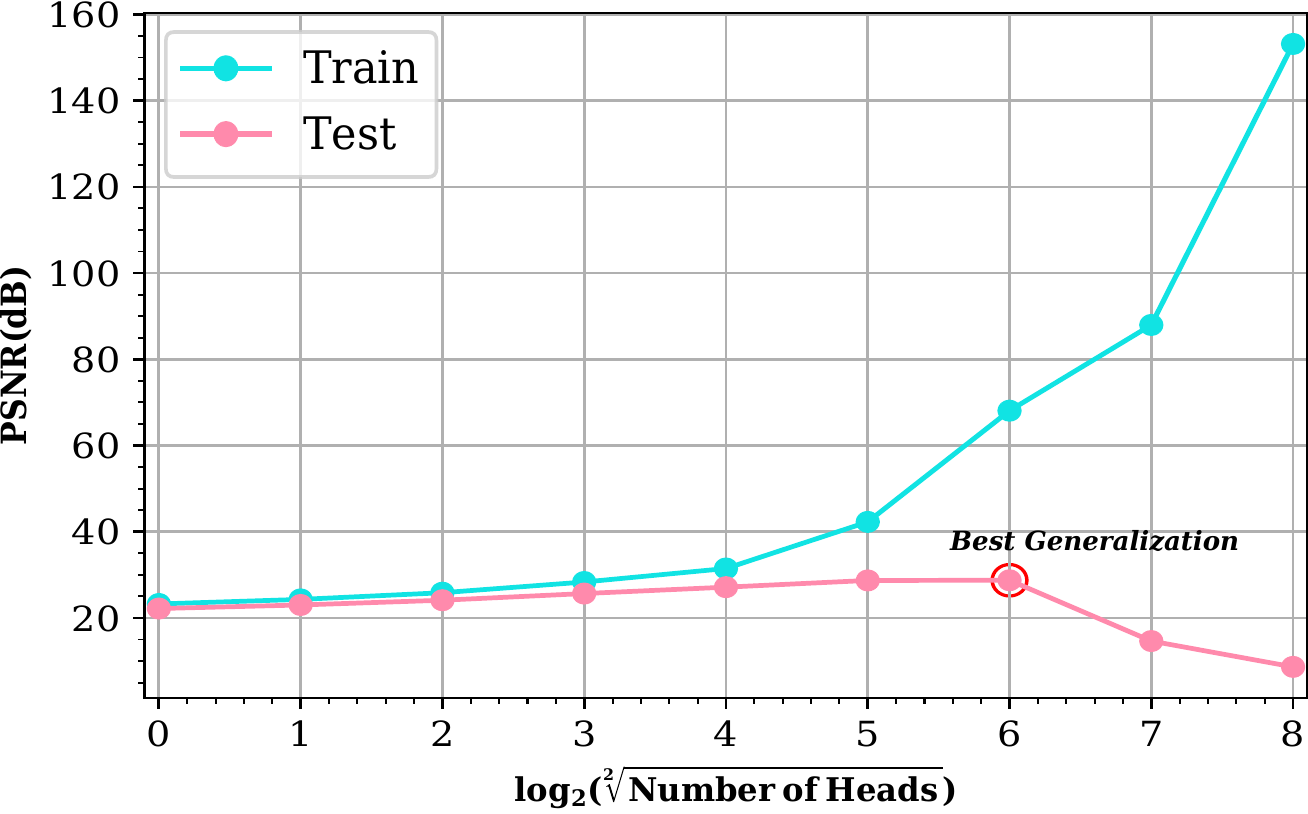}
				\caption{\small Generalization ability of the proposed method in different numbers of heads}
				\label{fig:generalization}
				
			\vspace{-1mm}
			\end{figure}
\subsubsection{Head}
One neuron is used for each rendering head $\tau_{l,k}(\cdot)$ to approximate the pixel values of the corresponding cell of the image.
The main advantage of using multiple heads to reconstruct different parts is the ability of the network to reconstruct several pixels in each forward pass, which leads to a significant reduction in model computations to reconstruct the whole image.

Alternatively, all of the single neurons can be considered as a fully connected layer, namely, the rendering multi-head layer, which is attached to the body network.
The main issue with a fully-connected rendering multi-head network is the the large number of required parameters, which is proportional to the number of heads connected to the model body.
In particular, the number of parameters grows significantly larger as the number of heads increase.
For example, if $256^2$ heads are used for a $256 \times 256$ grayscale image and the last layer of the model body has $256$ neurons, then approximately $16.7$ million (M) parameters are required.

To solve the problem, we use a sparse layer instead of the fully-connected one. In the sparse layer, the number of trainable parameters is significantly reduced due to omission of most of the connections with the previous layer.
% Since each neuron reconstructs a specific part of the image, it must be connected to the previous layer. 
Let us denote the number of connections of each neuron with the previous layer by $\alpha$.
Indeed, the minimum value of $\alpha$ is $1$, and by increasing this parameter to the number of neurons in the previous layer, the sparse layer becomes a fully-connected layer. 
%The rendering stage encompasses several heads each of which is responsible for generating pixel values of its corresponding pre-selected block image.
%The head generates several pixel values in each forward pass.

\subsection{Model Configuration}
\label{sec:conf}
To further analyse the model and tuning the hyper-parameters, a base body network with $4$-hidden layers and $256$ neurons for each layer is used. Each rendering head (output neuron) is connected to the body network with a partially-connected linear layer (sparse layer) that has only $\alpha$ connections to the output layer of the body. The active connections are randomly selected with a uniform distribution at the beginning of the training and do not change. For all the experiments and simulations, the weights and biases of the networks are initialized with uniform distribution, and training is performed for 2000 epochs.

\begin{figure}[t]
% \vspace{-2em}
			\centering
			\includegraphics[scale = 0.57]{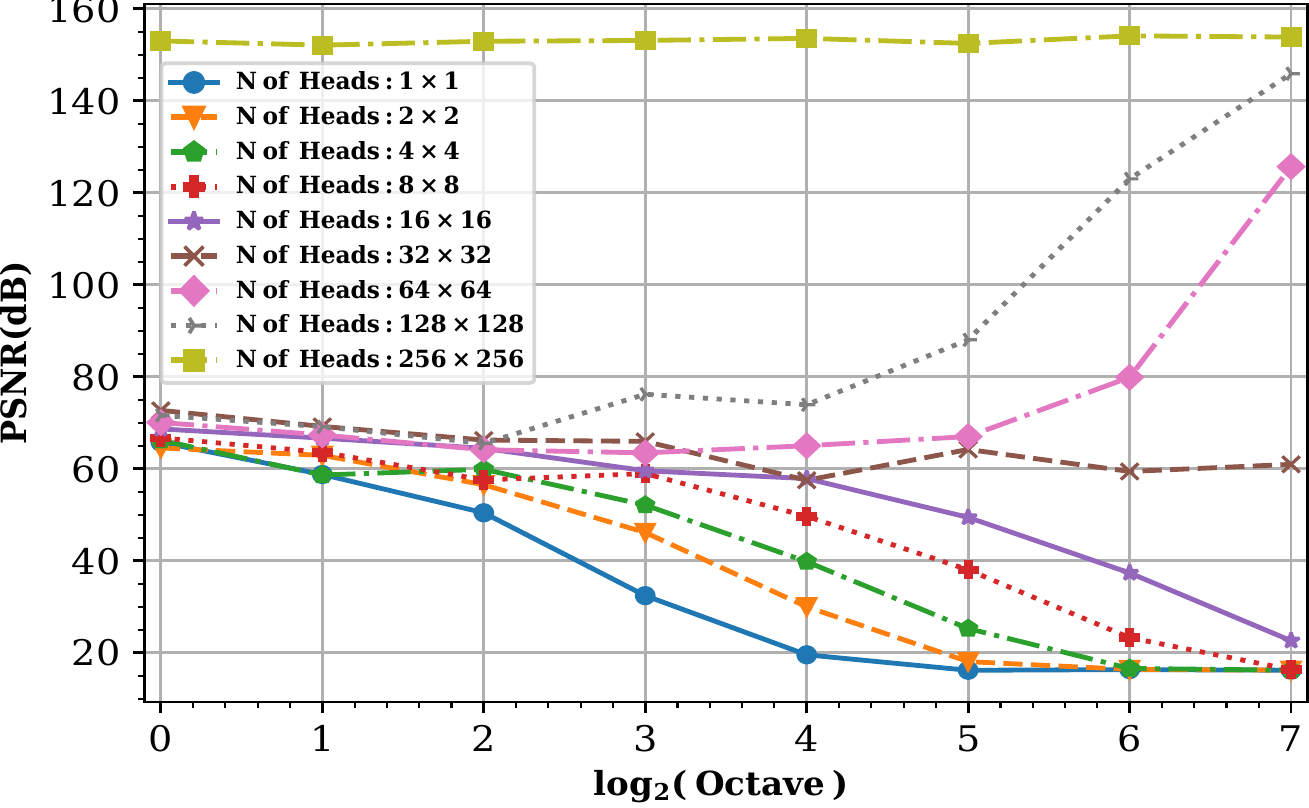}
				\caption{\small Spectral bias in the proposed method with different number of heads}
				\label{fig:frequency_bias}
				
			\vspace{-1mm}
			\end{figure}

\subsection{Spectral Bias}
Here, we present experimental results to show the effect of number of heads on the spectral bias of the model. Specifically, we use 2D Perlin noise \cite{perlin2002improving} as the target image. Perlin noise is a pseudo-random pattern of float values generated across an N-dimensional plane which allows for controlled high-frequency features in each dimension using a parameter, namely, Octave. In this experiment, Perlin noises with different frequencies are generated in two dimensions with a size of $256 \times 256$, shown in Fig.~\ref{fig:perlin_niose}. 

%We trained the proposed model with the different number of heads on this set of synthetic noises. The model hyperparameters, including the number of neurons and the $\alpha$ parameter, were selected so that the number of model parameters was the same in all comparisons. 
We trained the proposed model with the different number of heads on this set of synthetic noises. The base model presented in Subsection~\ref{sec:conf} is used for this experiment with $\alpha = 32$. 
In Fig.~\ref{fig:frequency_bias}, peak signal-to-noise ratio (PSNR) of the reconstructed image with different network configurations for various Octaves of Perline noise are presented.

Note that, the higher the Octave the higher the frequency of the target image. The solid blue curve which corresponds to the network with 1 head (which is in essence the regular ReLU network) clearly illustrates the spectral bias of such ReLU networks. However, as the number of heads increase, the model can more effectively reconstruct higher frequencies. In particular, by increasing the number of heads, each rendering unit (head) reconstructs a smaller area of the image. The amount of variation in the neighbouring pixels decreases dramatically with decreasing size of each cell, and in practice, each head reconstructs a low-frequency signal. 

Since each rendering head represents a hyperplane, if the number of model heads is equal to the number of image pixels, the hyperplane associated with each head can approximate that single pixel perfectly. In fact, for the case of $256^2$ where each head only reconstructs one pixel and the whole image is reconstructed with one forward pass, the error-free approximation of the network is shown in Fig.~\ref{fig:frequency_bias} regardless of frequency variations. 

\begin{figure*}[t]
			\centering
			\includegraphics[width=170mm,scale=1.0]{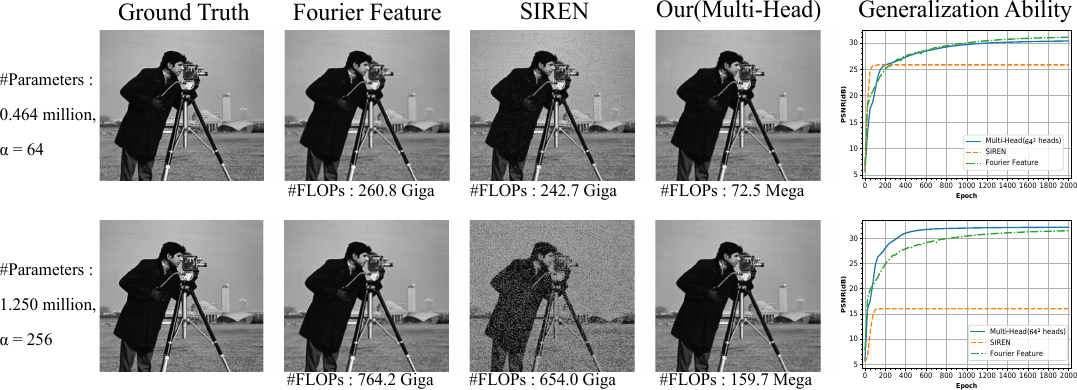}
				\caption{Comparison of the proposed method($64^2$ heads) with the SIREN and Fourier feature in terms of generalization ability and computational cost. The computational cost is calculated for the evaluation phase (image's size of $512 \times 512$).}
				\label{fig:comparison}
			
			\vspace{1mm}
			\end{figure*}

\subsection{Generalization ability}
% One of the significant advantages of neural networks is their ability to generalize. Generalization ability means that a trained model could predict data from the same distribution as the learning data that it has never seen before. In real-world applications, developers typically have only a part of all possible data points of the main distribution for the training of a neural network. 
In INR tasks, only memorizing the training data is not enough. Specifically, in addition to the ability of the proposed model to alleviate the frequency bias of ReLU networks, the generalization ability must be also considered. For example, in neural radiance fields (NERF)~\cite{mildenhall2020nerf}, continuous scenes are represented as 5D neural radiance fields, parameterized as MLP networks; if the model fails to create new scenes, it becomes practically useless.

Consequently, we presented an experiment to evaluate the generalization ability of the proposed model. For this experiment, we first resizes a $512 \times 512$ grayscale image to a $256 \times 256$ image. Then, the proposed model with different heads is trained on the resized image. Finally, the trained model's performance is evaluated on the original $512 \times 512$ image. Fig.~\ref{fig:generalization} shows the results of this experiment. We can observe that although the model's performance on the training data improves by increasing the number of heads, from the number of $64^2$ heads on-wards, we encounter a decrease in the generalization ability. The base model used in this experiment has $256^2$ heads and the $\alpha = 32$. Consequently, in the reset of the paper, we focus on a model with $64^2$ as it exhibits the best generalization characteristics.

\section{Experimental Results}
In this section, in order to compare our method with the state-of-the-art methods, we investigate the accuracy of different models with the same number of parameters. We also report number of floating point operations (FLOPs) for training each model.
%In order to make a fair comparison, the number of training parameters and the number of training epochs is considered the same.
Fig.~\ref{fig:comparison} illustrates the reconstructed image of the proposed model as well as SIREN~\cite{sitzmann2020implicit} and Fourier feature~\cite{Fourier_features}.
To ensure a fair comparison, SIREN and Fourier feature are configured (the best possible configurations are selected based on \cite{sitzmann2020implicit,Fourier_features}) to have the same number of parameter as the multi-head network with $\alpha=64$ and $256$, i.e., $0.464$ and $1.250$ million, respectively.
With $64^2$ heads, the proposed model with $0.464M$ parameters requires $\mathbf{3347}$ times fewer FLOPs compared to competitors to reconstruct the whole image, which shows its outstanding performance. It should be noted that by increasing the model's size from $0.464M$ to $1.250M$, the reduction of the required FLOPs becomes more pronounced ($\mathbf{4095}$ times). 
% (Fig.~\ref{fig:flops})
% Less computational cost in addition to evaluation time reduces model training time. 

%\begin{figure}[t]
%			\centering
%			\includegraphics[scale = 0.57]{flops_log.pdf}
%				\caption{FLOPs}
%				\label{fig:flops}
%			\end{figure}
			
\section{conclusion}
In this paper, a novel multi-head INR was proposed to improve the spectral bias of ReLU networks. In particular, tackle this issue by exporting the local structure of the signals. Specifically, an MLP is used as the body to capture the global features and several heads are used to reconstruct local features. The proposed structure requires considerably less computational cost while achieving superior or similar results to the state-of-the-art methods. 

%\vfill\pagebreak

\bibliographystyle{IEEEtran}
\bibliography{IEEEabrv,Biblo}

\end{document}